\documentclass[lettersize,journal]{IEEEtran}
\usepackage{amsmath,amsfonts}
\usepackage{algorithmic}
\usepackage{algorithm}
\usepackage{array}
\usepackage[caption=false,font=normalsize,labelfont=sf,textfont=sf]{subfig}
\usepackage{textcomp}
\usepackage{stfloats}
\usepackage{url}
\usepackage{verbatim}
\usepackage{graphicx}
\usepackage{cite}
\usepackage{multirow}
\begin{document}

\title{Transferring Modality-Aware Pedestrian Attentive Learning for Visible-Infrared Person Re-identification}

\author{Yuwei Guo, Wenhao Zhang, Licheng Jiao, \IEEEmembership{Fellow, IEEE}, Shuang Wang, Shuo Wang, and Fang Liu}

\markboth{Journal of \LaTeX\ Class Files,~Vol.~14, No.~8, August~2021}%
{Shell \MakeLowercase{\textit{et al.}}: A Sample Article Using IEEEtran.cls for IEEE Journals}


\maketitle

\begin{abstract}
Visible-infrared person re-identification (VI-ReID) aims to search the same pedestrian of interest across visible and infrared modalities. Existing models mainly focus on compensating for modality-specific information to reduce modality variation. However, these methods often lead to a higher computational overhead and may introduce interfering information when generating the corresponding images or features. To address this issue, it is critical to leverage pedestrian-attentive features and learn modality-complete and -consistent representation. In this paper, a novel Transferring Modality-Aware Pedestrian Attentive Learning (TMPA) model is proposed, focusing on the pedestrian regions to efficiently compensate for missing modality-specific features. Specifically, we propose a region-based data augmentation module PedMix to enhance pedestrian region coherence by mixing the corresponding regions from different modalities. A lightweight hybrid compensation module, i.e., the Modality Feature Transfer (MFT), is devised to integrate cross attention and convolution networks to fully explore the discriminative modality-complete features with minimal computational overhead. Extensive experiments conducted on the benchmark SYSU-MM01 and RegDB datasets demonstrated the effectiveness of our proposed TMPA model.
\end{abstract}


\section{Introduction}
Pedestrian re-identification (ReID) aims at matching the pedestrians from an image gallery captured by non-overlapping camera views. Most existing ReID models focus on visible-modality pedestrian matching captured by visible cameras, resulting to a performance degradation under poor illumination conditions (e.g., at nighttime). Infrared cameras are therefore introduced for such conditions, motivating increasing research on visible-infrared person re-identification (VI-ReID) recently.

The main challenge of modality variation in VI-ReID includes inter- and intra- variation (e.g., different viewpoints and color discrepancy) \cite{ye2018hierarchical}. To tackle the issue, most existing VI-ReID models \cite{jia2020similarity,park2021learning,chen2022structure} have been proposed to learn modality-shared feature representations between visible and infrared modalities. However, these approaches focus solely on the most discriminative modality-shared characteristics while overlooking modality-specific features which could significantly enhance pedestrian re-identification. 

In VI-ReID, modality-specific information compensation based models have become the prevailing paradigms. Typically, they utilize generative adversarial networks (GANs) \cite{wang2019rgb,wang2020cross,9399508} or feature projection \cite{lu2020cross,jiang2022cross,li2022visible} to generate the opposite modal images (or features). Then, these images (or features) are combined with the original ones to compensate for the missing modality-specific information. However, there are two issues in these paradigms. Firstly, these models incur excessive computational costs due to the numerous magnitude parameters of networks. Secondly, these models may suffer from interfering information. For example, one-to-many mapping is used to generate a visible image from the infrared modality, which results in color inconsistency and distinctly degrades the performance, as demonstrated in \cite{liu2022revisiting}. 

\begin{figure}[t]
	
	\centering
	\includegraphics[scale=0.2]{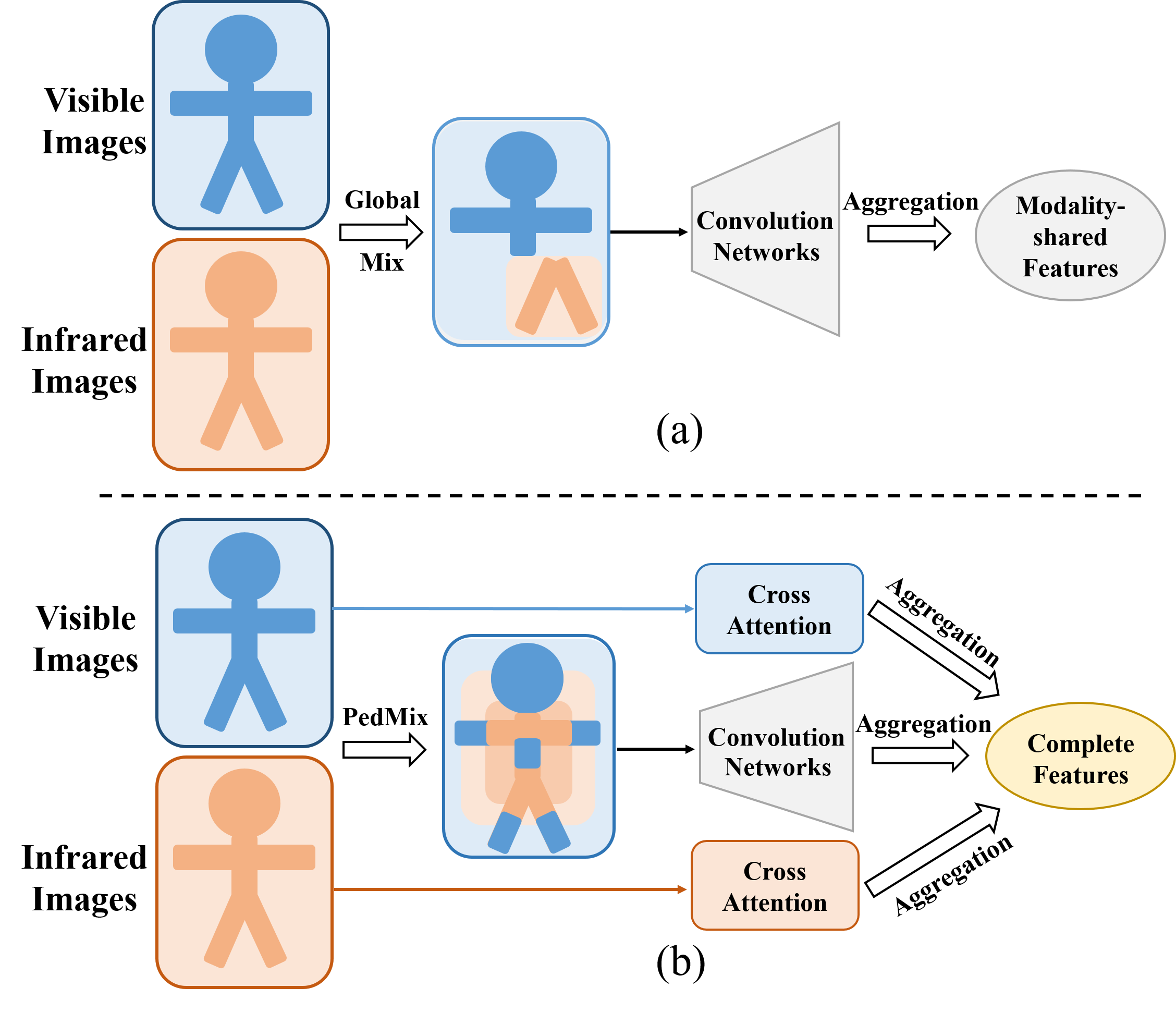}
	\caption{Comparison of the existing VI-ReID models and our TMPA model. (a) Existing models use global mix strategies to exploit modality-shared features. (b) Our TMPA model uses region-based PedMix and combines the cross attention with convolution to compensate for complete features.}
	\label{intro}
	\vspace{-10pt}
\end{figure}

Meanwhile, data augmentations are usually employed to improve model generalization and alleviate data scarcity scenarios. There are three mainstream augmentations in VI-ReID models. Firstly, channel augmentation\cite{ye2021channel} generates color-irrelevant images by randomly filling the RGB channel of visible images, thereby reducing the color discrepancy between two modalities. Secondly, global mix strategy \cite{qian2023visible,kim2023partmix} expands the training samples for consistent modality-shared features, as shown in Fig. 1(a). Thirdly, some works \cite{hua2023cross} combine these two data augmentations aiming for a better performance. However, the pedestrian and the context image parts usually contribute equally to augment the images in these approaches, which may generate improper images degrading model performance, e.g., random global mixing of two images may result in ambiguous or unnatural patterns.

Considering these challenges in data augmentation and modality-specific compensation, we propose a Transferring Modality-Aware Pedestrian Attentive Learning (TMPA) model for VI-ReID as shown in Fig. 1(b), including a region-based data augmentation module PedMix, a Modality Feature Extraction module (MFE), and a hybrid modality compensation module Modality Feature Transfer (MFT). Specifically, in the PedMix module, we aim to redirect the model's focus towards entire pedestrian areas rather than the context parts. Thus, the images are first segmented into three regions according to the pedestrian position. The mixed images are generated based on different regions. Compared with other mix-based data augmentations, our proposed PedMix module retains the coherence of the pedestrian regions and reduces the generation of ambiguous and unnatural images. Then, the mixed and the original images are decomposed into modality-shared and modality-specific features by the MFE module. Moreover, considering that existing modality-specific compensation tends to introduce numerous parameters and ID-irrelevant information, we propose a lightweight hybrid modality compensation module by cross attention and convolution networks. Cross attention applies a weighted average operation based on the context of different features, dynamically generating attention weights between the cross modality pixel pairs. Meanwhile, convolution networks leverage an aggregation function within a local receptive field to generate discriminative features in images. In our proposed MFT, by $1\times1$ convolutions, cross attention and convolution networks are elegantly integrated. This integration allows us to compensate for modality-specific information at minimal computational overhead and benefit our model from both paradigms. Finally, these compensatory features are fused with the modality-shared ones for modality-complete and –consistent representation.

The main contributions can be summarized as follows:

\begin{itemize}
	\item We propose a lightweight modality-specific compensation model TMPA for VI-ReID, which learns pedestrian attentive and modality-consistent feature representation. To the best of our knowledge, this is the first work that explores the integration of cross attention and convolution to compensate for the missing modality-specific information.
	\item We propose a data augmentation module PedMix tailored for VI-ReID, which mixes the visible and infrared images based on the pedestrian regions to preserve regional coherence and reduce the generation of ambiguous or unnatural images.
	\item Extensive experimental results on two standard benchmarks demonstrate that our TMPA model performs favorably against the state-of-the-art VI-ReID models with minimal computational overhead due to the efficient utilization of pedestrian features in both modalities.
\end{itemize}

\section{Related Works}
The existing VI-ReID models are usually classified into two categories, i.e., modality-shared feature learning based ones and modality-specific information compensation based ones.
\subsection{Modality-Shared Feature Learning}
The modality-shared feature learning based models usually map the features from different modalities into a shared space to learn discriminative modality-shared features. Wu \emph{et al.} \cite{wu2017rgb} first proposed a deep zero-padding network to embed the visible and infrared features into shared ones and further created a pioneering benchmark SYSU-MM01. Jia \emph{et al.} \cite{jia2020similarity} explored the intra-modality similarities and designed an effective SIM model to match the hard positive samples. Ye \emph{et al.} \cite{ye2021deep} proposed a two-stream model to learn the shareable features from two modalities. Following this, Gao \emph{et al.}. \cite{gao2021mso} proposed an MSO network to extract the shareable features from the single-modality space and common space. Then Sun \emph{et al.} \cite{sun2022not} further employed contrastive learning to match the positive samples at the pixel level, thus significantly reducing modal discrepancy.

\begin{figure*}[h]
	\centering
	\includegraphics[scale=0.1]{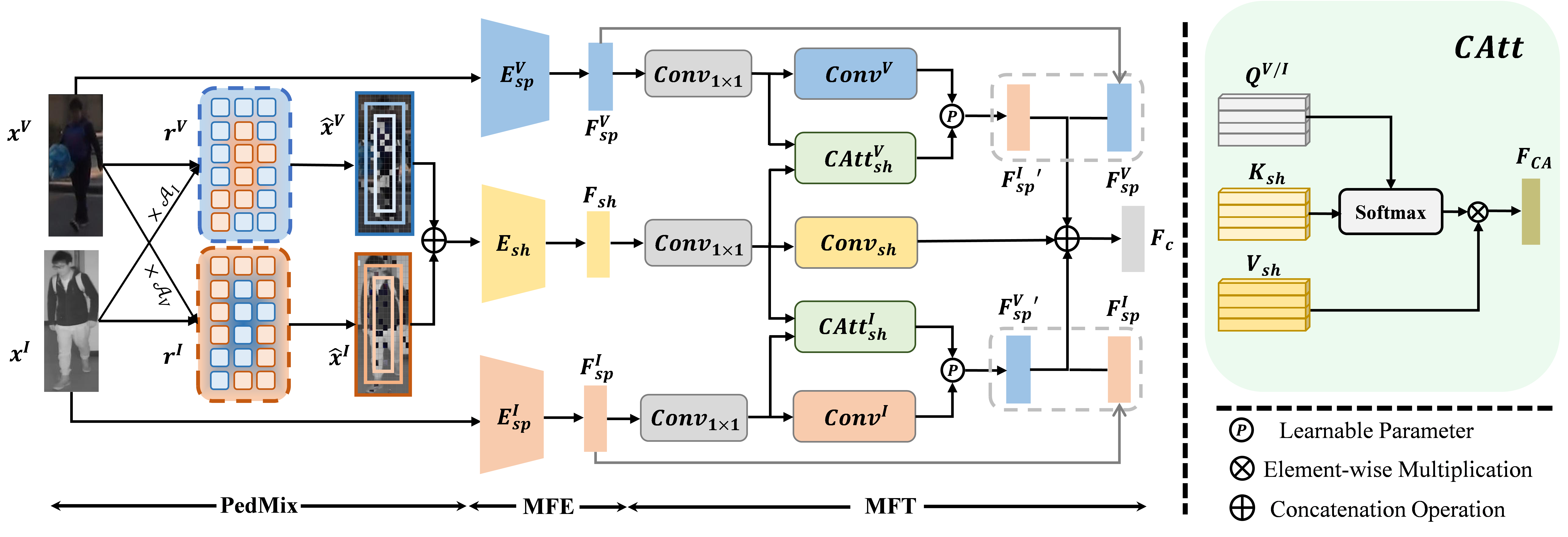}
	\caption{The overall architecture of our proposed TMPA model consists of three modules: a region-based data augmentation module PedMix, a modality feature extraction (MFE) module, and a modality feautre transfer (MFT) module.}
	\label{method}
	\vspace{-10pt}
\end{figure*}

\subsection{Modality-Specific Information Compensation}
Different from the modality-shared feature learning based models, the modality-specific information compensation based models exploit the modality-specific features to make up the missing ones from the existing modalities. These models can be categorized into adversarial generation based ones and feature transfer based ones \cite{lu2020cross}. 
The adversarial generation based models aim to create the missing modality images or features by GANs. Wang \emph{et al.} \cite{wang2020cross} proposed a paired-images generation method for the opposite modal images and then align the features at the set level and instance level. Liu \emph{et al.} \cite{liu2022revisiting} presented a two-stage GAN-based TSME model, first generating the high-quality images and then extracting the discriminative features by an attention mechanism. Zhang \emph{et al.} \cite{zhang2022fmcnet} employed GAN in the feature compensation after modality-specific features extraction. Differently, feature transfer based models aim to transfer the modality-specific features into the missing modality features, thereby achieving feature compensation. Lu \emph{et al.} \cite{lu2020cross} transferred the modality-specific features after modeling the affinities among different samples. Jiang \emph{et al.} \cite{jiang2022cross} proposed a modality-level alignment model CMT then by adaptive modulation the query and gallery instances are aligned in the same class. Li \emph{et al.} \cite{li2022visible} exploited missing modality information completion with a memory network. Then after a unified feature alignment, a consistent feature can be generated, alleviating the modality discrepancy. 

However, the existing models usually treat the complementary units and the feature processing units separately, ignoring the underlying relationship between them. This relationship may help the models learn discriminative features. Accordingly, we explore the relationship between them by integrating the complementary features with the semantic features to improve the discrimination of features while reducing the computational overhead.
\subsection{Data Augmentation}
Data augmentations are widely adopted to alleviate data scarcity and reduce overfitting during model training \cite{yun2019cutmix,zhang2017mixup,qian2023visible}. Among the augmentation methods, Ye \emph{et al.} \cite{ye2021channel} designed a CAJL model to randomly mix the channels of the entire visible images. Yang \emph{et al.} \cite{yang2022learning} proposed a DART model to mix globally cross modality images of the same pedestrian as new training samples. Ling \emph{et al.} \cite{ling2020class} proposed a class-aware modality mix strategy to generate global information about the visible and infrared modalities, closing the modality gap at the pixel level. Kim \emph{et al.} \cite{kim2023partmix} first extracted the part-level features from the whole images and then designed a positive and negative sample mixing strategy. 

Moreover, some works have aimed to enhance network generalization ability by emphasizing salient regions within mixed images. Attentive cutmix and puzzle mix \cite{walawalkar2020attentive,kim2020puzzle} utilize a pre-trained network to initially extract regions of interest that the model focuses on. Then the images are mixed based on these regions. TransMix \cite{chen2022transmix} weights Attention maps from different positions to alleviate discrepancies between input and label spaces.

While these models alleviate the generation of unnatural images to some extent, they ignore the importance of specific pedestrian regions. In VI-ReID tasks, pedestrian areas often occupy a significant portion of the image. Utilizing pre-trained networks or CAM algorithms fails to reasonably mix all salient parts of pedestrians and cannot facilitate end-to-end training. Additionally, these methods divide the continuous images or features into small parts for mixing without considering regional coherence. Differently, we propose a pedestrian attentive data augmentation module PedMix to mix the visible and infrared images based on the pedestrian positions while reserving the regional coherence and reducing the generation of unnatural images.

\subsection{Attention on Convolution}
To address the limitation of local interactions in convolutional networks, many works employ attention mechanisms as fundamental building blocks in vision tasks. Specifically, channel attention mechanisms such as Squeeze-and-Excitation (SE) \cite{hu2018squeeze} recalibrate the weight of each channel. Spatial attention mechanisms such as Spatial Transformer Networks (STN) \cite{jaderberg2015spatial}, correct affine-transformed targets, imparting spatial invariance to the network. Additionally, Convolutional Block Attention Module (CBAM) \cite{woo2018cbam} combines spatial and channel attention to reweight both channels and spatial positions to enhance the refinement of feature maps.

With Transformers demonstrating remarkable expressiveness in capturing long-range dependencies, several models aim to use self-attention in convolution networks. AA-Resnet \cite{bello2019attention} enriches certain convolutional layers by integrating attention maps from an independent self-attention pipeline. BoTNet \cite{srinivas2021bottleneck} replaces convolutions with self-attention modules in later stages of the model. DFLN-ViT \cite{zhao2022spatial} leverages discriminative feature learning through spatial and channel awareness.

However, these models treat convolution networks and attention as distinct parts, ignoring their potential relationship. In this paper, we propose a hybrid modality compensation module that elegantly integrates convolution and attention through carefully designed $1\times1$ convolutions. Consequently, this fusion reduces the introduction of ID-irrelevant information during modality-specific information compensation, while maintaining minimal computational overhead.

\section{Methods}

In this section, we introduce the Transferring Modality-Aware Pedestrian Attentive Learning (TMPA) model in detail. Firstly, we present the architecture of our model and the problem formulation. Then, we introduce three new modules in the TMPA model, PedMix, MFE, and MFT. Finally, we describe the loss functions to regularize our model.
\subsection{Overview}
As shown in Fig. 2, the TMPA model mainly consists of three modules: the data augmentation module PedMix, the Modality Feature Extraction module (MFE), and the Modality Feature Transfer module (MFT). The PedMix module is designed to mix visible and infrared images, generating clear and natural images. Then, these images are decomposed into modality-shared features and modality-specific features by the MFE module. Finally, a lightweight hybrid modality compensation module MFT is proposed, which integrates the cross attention mechanism with the convolution networks to generate the missing modality-specific features from the existing ones for modality-complete representation. We further employ several loss functions to regularize our model, including the proposed modality-shared-specific (MSS) loss function, modality-specific identity (MSI) loss function, as well as the commonly used weighted regularization triplet (WRT) loss function and identity classification (ID) loss function. Implementation details of our model will be discussed in the following subsections.

For the training data, let $D=\left\{X^V, X^I\right\}$ denote the labeled training dataset, where $V$ represents the visible modality and $I$ represents the infrared modality. The training dataset includes a visible image subset $X^V=\left\{x_{p, n}^V, p=1, \ldots, P ; n=1, \ldots, N\right\}$ and an infrared subset $X^I=\left\{x_{p, n}^I, p=1, \ldots, P ; n=1, \ldots, N\right\}$ , where $P$ and $N$ denote the identities and sample numbers, respectively.

\subsection{PedMix Module}
As we discussed earlier, the existing data augmentations using global mix may generate ambiguous or unnatural images. Moreover, the distribution characteristic of pedestrians located in the center of the image is not fully exploited, thus resulting into a model degraded performance. Based on this, we propose a novel data augmentation module PedMix tailored to the pedestrian regions, generating more natural images. 

For simplicity, only the visible image subset $X^V$ is described as an example. Specifically, the visible images $x^V \in X^V$ are first segmented into non-overlapping $T$ image patches $\tilde{x}^V$ with the same size and shape. Then, we employ two central rectangular boxes $\Phi_k, k \in[1,2]$ to organize these patches into three regions $r_i^V, i \in[\mathrm{C}, \mathrm{S}, \mathrm{O}]$ ,where $\mathrm{C}, \mathrm{S}$ and $\mathrm{O}$ denote the center, sub-center, and outer region, respectively. For VI-ReID, pedestrians deserve more attention. Therefore, we allocate the mask-out ratio $\mathcal{A}_i$ based on the proportion of pedestrians in different regions. When $\mathcal{A}_i=0$, the images only constitute the single-modality image. With $\mathcal{A}_i$ increasing, more inter-modality image patches are selected to generate the mixed image. According to each mask-out ratio, the image mask $M_i$ can be generated in each region. Formally,

\begin{equation}
	M_i=\mathcal{A}_i \times r_i^V.
\end{equation}

Finally, we apply these region-based masks $M_i$ to mix visible and infrared images of the same region with the same identity. The mixed visible image $\hat{x}^V$ can be represented as follows: 

\begin{equation}
	\hat{x}^V=\sum_i M_i \otimes x^V+\sum_i\left(1-M_i\right) \otimes x^I,
\end{equation}


\noindent where $\otimes$ denotes the element-wise multiplication. Similarly, we may obtain mixed infrared images $\hat{x}^I$. We concatenate the mixed visible and infrared images as the final mixed image $\hat{x}$, i.e,

\begin{equation}
	\hat{x} = \mathrm{Cat}\left(\hat{x}^V, \hat{x}^I\right),
\end{equation}


\noindent where $\mathrm{Cat}\left(*\right)$ denotes the concatenation operation. Notably, PMCM \cite{qian2023visible} also tries to mix the images with the segmented image patches. It first randomly generates the patch-mixed images based on the entire image. Then, the mixed images are sent into an auxiliary path to reduce the modality discrepancy. Different from PMCM, our proposed PedMix makes efficient use of the pedestrian distribution characteristic and retains the coherence of the pedestrian regions.

\subsection{Modality Feature Extraction Module}
In this subsection, we design a Modality Feature Extraction (MFE) module, including three feature extractors $E_{s h}(*)$, $E_{s p}^V(*)$ and $E_{s p}^I(*)$ to decompose the mixed image $\hat{x}$, the original visible image $x^V$ and the original infrared image
$x^I$ into the modality-shared $F_{sh}$ and modality-specific features $F_{sp}^V$, $F_{sp}^I$, respectively. The parameters of these feature extractors are mutually independent.

To regularize the modality-shared feature $F_{sh}$ and modality-specific features   $F_{sp}^V$, $F_{sp}^I$ simultaneously, a modality-shared-specific (MSS) loss is proposed, which constrains the feature generation at the visible-infrared level and shared-specific level. At the visible-infrared level, the visible modality-specific feature $F_{sp}^V$ is pushed away from the infrared modality-specific feature $F_{sp}^I$ to keep the modal specificity. At the shared-specific level, we concatenate the visible and infrared modality-specific features as the modality-specific feature $F_{sp}$. The modality-shared feature $F_{sh}$ is pushed away from the modality-specific feature $F_{sp}$. To this end, the MSS loss consists of two items, including a Visible-Infrared specific feature distance $d_{sp}$, and a Shared-Specific feature distance $d_{ss}$, which can be calculated by

\begin{equation}
	L_{MSS}=\sum_{p=1}^P \max \left(\rho-d_{s p}^p-d_{s s}^p, 0\right),
\end{equation}
\begin{equation}
	F_{sp}= \mathrm{Cat}(F_{sp}^V, F_{sp}^I),
	\vspace{-15pt}
\end{equation}

\begin{equation}
	\begin{split}
		d_{s p}=\frac{1}{N} \sum_{i=1}^N\left\|F_{s p}^V-F_{s p}^I\right\|_2, \\
		d_{s s}=\frac{1}{N} \sum_{i=1}^N\left\|F_{s h}-F_{s p}\right\|_2.
	\end{split}
\end{equation}

Here,  $\left\|*\right\|_2$ represents the operation of Euclidean distance and $\rho$ denotes the corresponding margin. In addition, to learn the identity-related features in each modality, the modality-specific identity loss (MSI) is defined as:

\begin{equation}
	\vspace{-5pt}
	\begin{split}
		L_{M S I}=-\frac{1}{N} \sum_{i=1}^N y_i \log \left(\mathcal{C}^V\left(F_{s p}^V\right)\right) \\ -\frac{1}{N} \sum_{i=1}^N y_i \log \left(\mathcal{C}^I\left(F_{s p}^I\right)\right).
	\end{split}
\end{equation}

Here, $y$ denotes the label of the training sample. $\mathcal{C}^V$ and $\mathcal{C}^I$ denote the visible and infrared modality-specific identity classifier, respectively, which is constructed by a Batch Normalization layer, a fully connected layer, and a softmax function. The final loss function used in the MFE module is calculated as 
\begin{equation}
	L_{M F E}=L_{M S I}+\lambda_1 L_{M S S}.
\end{equation}

\noindent where $\lambda_1$ denotes the weights of the proposed loss function $L_{MSS}$.
\subsection{Modality Feature Transfer  Module}
In the previous subsections, we employ the proposed PedMix to achieve a reasonable model generalization, allowing the model to focus on the entire region of pedestrians. And then the modality-shared and -specific features (i.e., $F_{sh}$, $F^V_{sp}$, $F^I_{sp}$) are decomposed by our proposed constraints. In this sub-section, we aim to compensate for modality-specific features with minimal computational overhead by exploring the relations between cross attention and convolution.

When performing a traditional $n \times n$ convolution operation, the convolutional kernel slides across the input feature map. We can consider the kernel's processing at each position as a series of $1 \times 1$ convolutional operations \cite{pan2022integration}, thus decomposing the output of an $n \times n$ convolution into a combination of $1 \times 1$ convolutional outputs at different positions. For example, for a $3 \times 3$ convolution, its convolution process can be represented as:

\begin{equation}
	O_{i, j}=\sum_{m=0}^2 \sum_{n=0}^2 I_{i+m, j+n} \times C^{Conv}_{m, n},
\end{equation}

\noindent where $I$ denotes the input feature map, $O$ denotes the output feature map and $ C^{Conv}_{m, n}$ represents the weights of the convolutional kernel at different positions. Based on this, we can decompose it into nine different positions of $1 \times 1$ convolutions. Each $1 \times 1$ convolution corresponds to the weights of the $3 \times 3$ convolution at different positions. Meanwhile, when performing cross attention, the input feature map is firstly projected into query, key, and value by $1 \times 1$ convolutions, i.e., $Q = C^{CA}_Q\times I $, $K = C^{CA}_K\times I $, $V = C^{CA}_V\times I $, where $C^{CA}$ denotes the projection weight. Then the attention weights can be computed by the cross calculation of query, key, and value, which are obtained from different modalities.

We can observe that both traditional convolution and cross attention share the part of $1 \times 1$ convolutions, followed by the aggregation of generated features. As shown in Fig. 2, to maintain minimal computational overhead for our proposed modality-specific information compensation method, we reuse the $1 \times 1$ convolutions and perform distinct aggregation operations on their outputs.

Specifically, our proposed MFT module comprises of two stages. In the first stage, we aim to transfer the modality-specific features between the visible and infrared modalities by two symmetric cross attention. In traditional cross attention, the two input data (e.g. semantic and visual information), are heterogeneous, so an asymmetric structure is designed to capture the correlation between them. Considering that the visible and infrared modality-specific features are highly homogeneous, the symmetric cross attention is designed to learn the complementary pedestrian attentive features. Specifically, the modality-shared feature $F_{sh}$ and modality-specific features $F_{sp}^V$, $F_{sp}^I$ are first mapped by $1\times1$ convolutions into projection matrices (i.e., key $K_{sh}$, value $V_{sh}$, and queries $Q_{sp}^V$, $Q_{sp}^I$ ). The key and value $K_{sh}$, $V_{sh}$ are shared by the specific queries $Q_{sp}^V$, $Q_{sp}^I$ to generate the intermediate features $F_{CA}^V$, $F_{CA}^I$. We use the following expressions to realize the cross attention operation:

\begin{equation}
	\begin{split}
		F_{C A}^V=\operatorname{softmax}\left(\frac{Q_{s p}^I K_{s h}^T}{\sqrt{d}}\right) V_{s h}, \\
		F_{C A}^I=\operatorname{softmax}\left(\frac{Q_{s p}^V K_{s h}^T}{\sqrt{d}}\right) V_{s h},
	\end{split}
\end{equation}

\noindent where $d$ is the dimension of the matrices. In addition, these projection matrices are reused in the convolution networks (i.e., $Conv^V$, $Conv^I$ and $Conv_{sh}$) to capture the semantic features $F_{Conv}^V$, $F_{Conv}^I$ and $F_{Conv}^{sh}$.

In the second stage, we aggregate these features to compensate for the missing modality-specific features. Specifically, we first combine the modality-specific semantic features $F_{Conv}^V$, $F_{Conv}^I$ with the intermediate features $F_{CA}^V$, $F_{CA}^I$:
\begin{equation}
	\begin{split}
		F_{s p}^V{ }^{\prime}= F_{C A}^V+ F_{C o n v}^V, \\
		F_{s p}^I{ }^{\prime}= F_{C A}^I+ F_{C o n v}^I.
	\end{split}
\end{equation}

Then, these generated features $F_{s p}^V{ }^{\prime}$ and $F_{s p}^I{ }^{\prime}$ are concatenated with the real ones $F_{s p}^V$ and $F_{s p}^I$ for unified feature representation. After that, the shared semantic features  $F_{Conv}^{sh}$ are added to generate modality-complete feature representation:

\begin{equation}
	\vspace{0pt}
	\begin{split}
		F_{C}= & \lambda_2 \times \mathrm{Cat}\left(F_{s p}^V{ }^{\prime}, F_{s p}^I\right) \\+ & \lambda_3 \times \mathrm{Cat}\left(F_{s p}^V, F_{s p}^I{ }^{\prime}\right)+F_{Conv}^{sh},
	\end{split}
\end{equation}


\noindent where $\lambda_2$ and $\lambda_3$ denote the weights of the concatenated features, respectively.

To maintain the the ID relevance of the generated features $F_{s p}^V{ }^{\prime}$ and $F_{s p}^I{ }^{\prime}$, the modality-transfer identity loss is defined as:
\begin{equation}
	\vspace{0pt}
	\begin{split}
		L_{MFT}=-\frac{1}{N} \sum_{i=1}^N y_i \log \left(\mathcal{C}^V\left(F_{s p}^V{ }^{\prime}\right)\right)\\
		-\frac{1}{N} \sum_{i=1}^N y_i \log \left(\mathcal{C}^I\left(F_{s p}^I{ }^{\prime}\right)\right).
	\end{split}
\end{equation}



\subsection{Loss Function Optimization}

Following the previous methods \cite{zhang2022fmcnet,liu2022revisiting,ye2021channel}, an identity classification (ID) loss function $L_{ID}$ \cite{zheng2017discriminatively} and a weighted regularization triplet (WRT) loss function $L_{WRT}$ \cite{ye2021channel} are employed.

The ID loss $L_{ID}$ is usually used to guarantee identity-aware discrimination, which encourages a specific ID-related feature representation. The $L_{ID}$ is denoted by:
\begin{equation}
	L_{I D}=-\frac{1}{N} \sum_{i=1}^N y_i \log \left(\mathcal{C}\left(F_C\right)\right),
\end{equation}
where $N$ denotes the number of identities, $ y_i $ denotes the ground-truth label, and 
$\mathcal{C}(*)$ denotes the pedestrian identity classifier.

The WRT loss $L_{WRT}$ aims to optimize the relative distance between all the positive and negative pairs. $L_{WRT}$ requires the samples to be drawn closer within the same ID and pushed away otherwise. The $ L_{WRT}$ is denoted by:

\begin{equation}
	\begin{gathered}
		W_{P, i} = \frac{\exp \left(d_{i i^{+}}\right)}{\sum_{i^{+}=1}^{N_{i^{+}}} \exp \left(d_{i i^{+}}\right)}, \\
		W_{N, i} = \frac{\exp \left(d_{i i^{-}}\right)}{\sum_{i^{-}=1}^{N_{i^{-}}} \exp \left(d_{i i^{-}}\right)}, \\
		L_{\text{WRT}} = \frac{1}{N} \sum_{i=1}^{N} \log \left[1+\exp \left(W_{P, i} d_{i i^{+}}-W_{N, i} d_{i i^{-}}\right)\right],
	\end{gathered}
\end{equation}

\noindent where, $N_{i^{+}}$ and $ N_{i^{-}}$ denote the positive samples and negative samples of the current sample, respectively. $ d_{i i^{+}}$ and $ d_{i i^{-}}$ denote the Euclidean distance between the positive pair and the negative pair, respectively. $W_{P, i}$ and $W_{N, i}$ denote the weight generated by the distance sets.

Here, we combine the $L_{ID}$ and $L_{WRT}$ as base loss $L_{base}$. Additionally, we combine the proposed $L_{MFE}$ and $L_{MFT}$ as modality feature loss $L_{MF}$. Thus, the total loss can be defined as
\begin{equation}
\vspace{0pt}	
L_{Total}= \alpha  L_{base}+\beta  L_{MF},
\end{equation}
where $\alpha $ and $\beta$ are the balance parameters. The detailed parameter analysis will be provided in the following section.

\begin{table*}[!h]
	\centering
	\caption{Comparison with state-of-the-art models on the SYSU-MM01 and RegDB datasets}
	\scalebox{0.95}{
		\begin{tabular}{c|c|cccccc|cccccc}
			\hline\rule{0pt}{10pt}
			\multirow{3}[3]{*}{Methods} & \multirow{3}[3]{*}{Venue} & \multicolumn{6}{c|}{SYSU-MM01} & \multicolumn{6}{c}{Regdb} \\
			\cline{3-14}\rule{0pt}{10pt}          &       & \multicolumn{3}{c}{All Search} & \multicolumn{3}{c|}{Indoor Search} & \multicolumn{3}{c}{Visible to Infrared} & \multicolumn{3}{c}{Infrared to Visible} \\
			\cline{3-14}\rule{0pt}{10pt}           &       & Rank-1 & Rank-10 & mAP   & Rank-1 & Rank-10 & mAP   & Rank-1 & Rank-10 & mAP   & Rank-1 & Rank-10 & mAP \\
			\hline\rule{0pt}{10pt}
			AlignGAN & ICCV-19 & 42.40  & 85.00  & 40.70  & 45.90  & 87.60  & 54.30  & -     & -     & -     & -     & -     & - \\
			JSIA  & AAAI-20 & 38.10  & 80.70  & 36.90  & 43.80  & 86.20  & 52.90  & -     & -     & -     & -     & -     & - \\
			cm-SSFT & CVPR-20 & 61.60  & 89.20  & 63.20  & 70.50  & 94.90  & 72.60  & 72.30 & -     & 72.90 & 71.00 & -     & 71.70 \\
			GECNet & TCSVT-22 & 53.37  & 89.86  & 51.83  & 60.60  & 94.29  & 62.89  & 82.33  & 92.72  & 78.45  & 78.93  & 91.99  & 75.58  \\
			RBDF  & TCYB-22 & 57.66  & 85.85  & 54.41  & -     & -     & -     & 79.80  & 93.59  & 76.71  & 76.21  & 90.77  & 73.92  \\
			CMDSF  & KBS-22 & 59.97  & 92.84  & 57.28  & -     & -     & -     & 84.75  & 92.16  & 77.91  & -  & -  & -  \\
			TSME  & TSCVT-22 & 64.23  & 95.19  & 61.21  & 64.80  & 96.92  & 71.53  & 87.35  & 97.10  & 76.94  & 86.14  & 96.39  & 75.70  \\
			\hline\rule{0pt}{10pt}
			SIM   & IJCAI-20 & 56.93  & -     & 60.88  & -     & -     & -     & 74.47  & -     & 75.29  & 75.24  & -     & 78.30  \\
			JCCL  & AAAI-21 & 57.20  & 94.30  & 59.30  & 66.60  & 98.80  & 74.70  & 78.80  & -     & 69.40  & 77.90  & -     & 69.40  \\
			FBP-AL & TNNLS-21 & 54.14  & 86.04  & 50.20  & -     & -     & -     & 73.98  & 89.71  & 68.24  & 70.05  & 89.22  & 66.61  \\
			LbA   & ICCV-21 & 55.41  & -     & 54.14  & 58.46  & -     & 66.33  & 74.17  & -     & 67.64  & 72.43  & -     & 65.46  \\
			PIC   & TIP-22 & 57.50  & -     & 55.10  & 60.40  & -     & 67.70  & 83.60  & -     & 79.60  & 79.50  & -     & 77.40  \\
			DART  & CVPR-22 & 68.72  & 96.39  & 66.29  & 72.52  & 97.84  & 78.17  & 83.60  & -     & 75.67  & 81.97  & -     & 73.78  \\
			SPOT  & TIP-22 & 65.34  & 92.73  & 62.25  & 69.42  & 96.22  & 74.63  & 80.35  & 93.48  & 72.46  & 79.37  & 92.79  & 72.26  \\
			DTRM  & TIFS-22 & 63.03  & 93.82  & 58.63  & 66.35  & 95.58  & 71.76  & 79.09  & 92.25  & 70.09  & 78.02  & 91.75  & 69.56  \\
			DML   & TCSVT-22 & 58.40  & 91.20  & 56.10  & 62.40  & 95.20  & 69.50  & 77.60  & -     & 84.30  & 77.00  & -     & 83.60  \\
			SPANet & TNNLS-23 & 65.74  & 92.98  & 60.83  & 71.60  & 96.60  & 80.05  & 76.31  & 91.02  & 68.00  & 70.15  & 85.24  & 63.77  \\
			CMTR  & TMM-23 & 65.45  & 94.47 & 62.90  & 71.46  & 97.16 & 76.67  & 88.11  & -     & 81.66  & 84.92  & -     & 80.79  \\
			PMT   & AAAI-23 & 67.53  & 95.39  & 64.98  & 71.66  & 96.73  & 76.52  & 84.83  & -     & 76.55  & 84.16  & -     & 75.13  \\
			\hline\rule{0pt}{10pt}
			TMPA(ours) & This paper & 68.84  & 95.61  & 66.42  & 72.65  & 97.15  & 76.76  & 88.69  & 97.27  & 81.95  & 86.31  & 96.80  & 79.77  \\
			\hline
	\end{tabular}}%
	\label{sota}%

\end{table*}%

\section{Experiment}
\subsection{Datasets and Metrics}
Our model is evaluated on two public VI-ReID datasets: SYSU-MM01 \cite{wu2017rgb} and RegDB \cite{nguyen2017person}.

\noindent \textbf{SYSU-MM01} is a challenging benchmark dataset for VI-ReID. The training set consists of 22,258 visible images and 11,909 infrared images of 395 identities. The testing set consists of 3803 infrared images of 96 identities, captured by four visible cameras and two infrared cameras. Meanwhile, there are two different search modes: all-search and indoor-search. The former is evaluated with indoor and outdoor images, while the latter only uses indoor images.

\noindent \textbf{RegDB} contains 412 identities, with 10 visible images and 10 infrared images per identity. The training and testing sets each consists of 2060 visible images and 2060 infrared images of 206 identities. The search modes in RegDB include Visible to Infrared mode and Infrared to Visible mode. The former takes the visible images as the query set and the infrared images as the gallery set, and vice versa.

Following the common used evaluation metrics for VI-ReID, the cumulative matching characteristics (CMC) and mean average precision (mAP) \cite{wu2017rgb,ye2018hierarchical} are adopted as evaluation metrics in this paper. Moreover, we conduct the experiments on four search modes, i.e., the all-search mode and the indoor-search on SYSU-MM01, the Visible to Infrared mode, and the Infrared to Visible mode on RegDB.
\subsection{Implementation Details}
Our TMPA model is implemented in the Pytorch framework with a single NVIDIA GeForce 3090 GPU for training. The input images $x^V$ and $x^I$ are resized to $288 \times 144$. The image patch $\tilde{x}$ is set to $12\times12$. Accordingly, in the PedMix module, each image is divided into 288 patches. The rectangular box $\Phi_1$ is empirically defined with dimensions equal to $2/3$ of the original image's length and $1/3$ of its width, while the box $\Phi_2$ has dimensions of $5/6$ of the length and $2/3$ of the width. To ensure accurate and reliable feature extraction, the backbone (i.e., ResNet-50) is pre-trained on the ImageNet. Moreover, the modality feature extractors (i.e., $E_{sh}$, $E_{s p}^V(*)$ and $E_{s p}^I(*)$) are constructed by the first three blocks of ResNet-50. In the MFT module, the convolution blocks (i.e., $Conv_{sh}$, $Conv_{sh}^V$ and $Conv_{sh}^I$) are constructed by the last two blocks of ResNet-50. In terms of hyper-parameters, the corresponding margin $\rho$ is empirically set to 0.65. The weights of loss functions (i.e., $\lambda_1$, $\lambda_2$ and $\lambda_3$) are empirically set to 0.2, 0.25, and 0.25. In the training stage, the batch size is set to 64, with 4 visible and 4 infrared images belonging to 8 identities. To optimize our model, we employ the stochastic gradient descent (SGD) optimizer with an initial learning rate of 0.1, which is decayed by a factor of 0.1 and 0.01 at the 20th and 50th epochs, respectively. Following \cite{ye2021deep}, the stride of the last convolutional layer in ResNet-50 is set to 1.

\subsection{Comparison with State-of-the-art Models}
In this subsection, as shown in Table \uppercase\expandafter{\romannumeral1}, we compare the TMPA model with the several state-of-the-art models to demonstrate the effectiveness. Specifically, the modality-shared feature learning models includes SIM \cite{jia2020similarity}, JCCL \cite{zhao2021joint}, FBP-AL\cite{wei2021flexible}, LbA \cite{park2021learning}, PIC \cite{zheng2022visible}, DART \cite{yang2022learning}, SPOT \cite{chen2022structure}, DTRM\cite{9665382} SPANet\cite{9525833}, DML\cite{zhang2022dual}, CMTR \cite{10130375}, and PMT \cite{lu2023learning}. The modality-specific information compensation models includes AlignGAN \cite{wang2019rgb}, JSIA \cite{wang2020cross}, cm-SSFT \cite{lu2020cross}, GECNet\cite{9399508}, RBDF\cite{wei2022rbdf}, CMDSF \cite{li2022cross} and TSME \cite{liu2022revisiting}.

On the SYSU-MM01 dataset, our proposed TMPA model achieves competitive performance compared with the state-of-the-art. Specifically, our model achieves 68.34\% Rank-1 accuracy, 66.42\% mAP under the all-search mode, and 72.65\% Rank-1 accuracy, 76.76\% mAP under the indoor-search mode, respectively. Moreover, we observe that compared with the GAN-based models (e.g., JSIA) our model surpasses by 30.24\% in Rank-1 accuracy and 29.52\% in mAP under all-search mode. Meanwhile, without the image-generation process, our model becomes more time-efficient and avoids the introduction of interfering information.

On the RegDB dataset, our proposed TMPA model outperforms all the compared models, achieving the best results on both search modes. Specifically, our model achieves the highest 88.69\% Rank-1 accuracy, 81.95\% mAP under the Visible to Infrared mode, and 86.31\% Rank-1 accuracy, 79.77\% mAP under the Infrared to Visible mode. TMPA makes a significant improvement of +3.11\% in mAP compared to the top-performing method TSME \cite{liu2022revisiting}. In conclusion, we can confirm the effectiveness of our proposed TMPA model from the above experiments.

\subsection{Ablation Study and Analysis}

In this subsection, we first conduct an ablation study to verify the impacts of different modules. Then, we further discuss the effectiveness of our proposed module under different settings. All experiments are conducted on the RegDB dataset under the Infrared to Visible mode.

\begin{table}[h]
	
	\centering
	\caption{Comparison with Different Modules of Our TMPA Model}
	\scalebox{0.9}{
		\begin{tabular}{c|cccc}
			\hline\rule{0pt}{10pt}
			Methods & Rank-1 & Rank-10 & Rank-20 & mAP \\
			\hline\rule{0pt}{10pt}
			Baseline (B) & 75.68 & 90.10  & 93.16 & 67.41 \\
			B+CA  & 76.83 & 91.38 & 93.69 & 68.27 \\
			B+CA+PedMix & 81.41 & 94.47 & 96.76 & 72.08 \\
			B+CA+PedMix+MFT & 86.31 & 96.80  & 98.06 & 79.77 \\
			\hline
	\end{tabular}}
	\label{xiaorong}%
\end{table}%

\begin{table}[h]
	\centering
	\caption{Comparison with Different Region Settings in PedMix}
	\scalebox{0.9}{
		\begin{tabular}{ccc|cccc}
			\hline\rule{0pt}{10pt}
			\makebox[0.045\textwidth][c]{center} & \makebox[0.045\textwidth][c]{sub-center} & \makebox[0.045\textwidth][c]{outer} &  Rank-1 & Rank-10 & Rank-20 & mAP \\
			\hline\rule{0pt}{10pt}
			&       &       & 76.83 & 91.38 & 93.69 & 68.27 \\
			$\checkmark$     &       &       & 77.82 & 92.09 & 94.90  & 69.42 \\
			$\checkmark$     & $\checkmark$     &       & 78.59 & 92.62 & 95.34 & 70.90 \\
			$\checkmark$     & $\checkmark$     & $\checkmark$    & 81.41 & 94.47 & 96.76 & 72.08 \\
			\hline
	\end{tabular}}%
	\label{region}%
\end{table}%

\begin{figure}[htbp]
	\centering
	\includegraphics[scale=0.42]{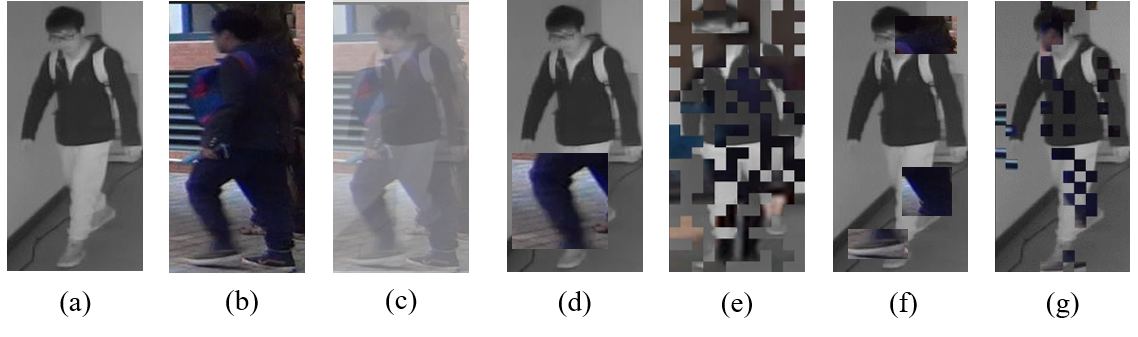}
	\centering
	\caption{Images from Different Mix Strategies. (a) Original visible image (b) Original infrared image (c) Mixup image (d) CutMix image (e) PatchMix image (f) Attentive CutMix image (g) PedMix image.}
	\label{A}
	\vspace{-5pt}
\end{figure}

\begin{table}[h]
	\centering
	\caption{Comparison with Different Mix Strategies and Baselines}
	\scalebox{0.9}{
		\begin{tabular}{c|cccc}
			\hline\rule{0pt}{10pt}
			Methods & Rank-1 & Rank-10 & Rank-20 & mAP \\
			\hline\rule{0pt}{10pt}
			B+Mixup & 76.07 & 91.94 & 94.66 & 69.50 \\
			B+CutMix & 75.92 & 90.40  & 93.08 & 68.85 \\
			B+PatchMix & 76.41 & 88.88 & 92.38 & 68.29 \\
			B+Attentive CutMix & 79.66 & 93.36 & 95.37 & 70.56 \\
			B+PedMix & 81.41 & 94.47 & 96.76 & 72.08 \\
			\hline\rule{0pt}{10pt}
			DART  & 81.97 & 95.05 & 97.04 & 73.78 \\
			DART+PedMix & 84.63 & 94.07 & 98.11 & 75.91 \\
			\hline
		\end{tabular}%
		\label{mix}}%

\end{table}%

\noindent \textbf{Effectiveness of proposed modules.} We first train a “Baseline (B)” model which only consists of the modality-shared feature extractor with ID loss function and WRT loss function as shown in Table \uppercase\expandafter{\romannumeral2} . Besides, we incorporate the random channel exchangeable augmentation (CA) \cite{ye2021channel} to “B+CA” to alleviate the color shift. Then, we evaluate the effectiveness of our proposed PedMix module. Compared with the “B+CA” model, when the PedMix is adopted, “B+CA+PedMix” significantly improves the performance by +4.58\% in Rank-1 accuracy and +3.81\% in mAP. This is because the region-based data augmentation mixes the corresponding regions between the two modalities, thereby retaining region coherence. Finally, we add the proposed MFT module to the model. The “B+CA+PedMix+MFT” model greatly outperforms “B+CA” by +9.48\% in Rank-1 accuracy and +11.5\% in mAP. This indicates that our MFT module can effectively transfer the modality-specific features to make up for the missing ones. Compared to the baseline model, which only considers the modality-shared features, our model fully exploits the pedestrian-related information in modality-specific features, boosting the performance.

\noindent \textbf{Analysis of PedMix.} As shown in Table \uppercase\expandafter{\romannumeral3}, we investigate the effectiveness of different combinations of center, sub-center, and outer regions in PedMix. When only the center region is selected, we can see the accuracy increase in all four metrics. After combining with the sub-center region, the performance can still obtain a slight improvement. When all the regions are combined, the performance is greatly improved compared with the baseline model. This indicates that pedestrians' proportions in each region significantly impact on the model's performance. Based on this, our PedMix helps the model to pay more attention to the pedestrian regions while retaining regional coherence.

Fig.3 and Table \uppercase\expandafter{\romannumeral4} show the comparison between our PedMix and other mix strategies. Notably, the CutMix \cite{yun2019cutmix} degrades the performance of the baseline. It may be attributed to the global random mixing, which interferes with the coherence of regions and consequently reduces the model's focus on pedestrians. On the other hand, PatchMix \cite{qian2023visible} shows a slight improvement in Rank-1 accuracy and mAP, however, it still inherits the limitation of the global mixing strategies. Moreover, we explore the effectiveness of our proposed PedMix in another baseline shown in “DART+PedMix”. The Rank-1 accuracy is improved by +2.66\%, and the mAP is improved by +2.13\% compared with “DART”. All evidence above shows that the proposed PedMix module effectively retains the coherence of the pedestrian regions and reduces the generation of ambiguous and unnatural images, achieving a better performance. In addition, attentive cutmix \cite{walawalkar2020attentive} is a mix strategy designed for image classification tasks. However, in VI-ReID, the resolution of pedestrian images is low, thus the pretrained backbone cannot accurately locate the entire pedestrian regions.

\begin{table}[t]
	\centering
	\caption{Parameter Comparison of Different Models}
	\scalebox{1.1}{
	\begin{tabular}{c|ccc}
		\hline\rule{0pt}{10pt}
		Methods & Params & Rank-1 & mAP \\
		\hline\rule{0pt}{10pt}
		AlignGAN & 82.0M & 56.30 & 53.40 \\
		JSIA  & 60.6M & 48.10 & 48.90 \\
		DART & 70.9M & 81.97 & 73.78 \\
		\hline\rule{0pt}{10pt}
		TMPA (w/o MFT) & 69.5M & 85.55 & 77.86 \\
		TMPA  & 48.3M & 86.31 & 79.77 \\
		\hline
	\end{tabular}}%
	\label{canshu}%
\end{table}%

\begin{table}[htbp]
	\centering
	\caption{Comparison of Different Attention Mechanism}
	\begin{tabular}{c|cccc}
		\hline\rule{0pt}{10pt}
		Methods & Rank-1 & Rank-10 & Rank-20 & mAP \\
		\hline\rule{0pt}{10pt}
		B\_DA+SENet & 81.62 & 95.63 & 97.22 & 73.31 \\
		B\_DA+STN & 82.33 & 94.96 & 96.84 & 74.57 \\
		B\_DA+CBMA & 84.27 & 96.32 & 97.65 & 77.22 \\
		B\_DA+MFT & 86.31 & 96.80 & 98.06 & 79.77 \\
		\hline
	\end{tabular}%
	\label{tab:addlabel}%
\end{table}%

\noindent \textbf{Analysis of MFT.} We conduct experiments to investigate the parameters and performance of different compensation methods. As shown in Table \uppercase\expandafter{\romannumeral5}, compared to the GAN-based models AlignGAN and JSIA, our TMPA achieves better performance with 48.3M Params. Meanwhile, compared to DART, our TMPA doesn't require additional computation at the label level, thus achieving lower parameters. Moreover, before activating the MFT module, (i.e., combine the cross attention after the convolution), we observe superior performance in terms of Rank-1 accuracy by +0.76\% and in mAP by +1.91\% while utilizing less computational cost.

Besides, due to the integration of convolutional operations and cross attention within our proposed MFT module, we further compare it with other attention mechanisms. As shown in Table \uppercase\expandafter{\romannumeral6}, where \text{B\_DA} denotes Baseline+CA+PedMix, we compare our proposed MFT with channel attention (i.e., SENet), spatial attention (i.e., STN), and the fusion of spatial and channel attention (i.e., CBMA). We can observe that employing solely spatial or temporal attention yields sight improvements. With CBMA, it is worth noting that mAP is enhanced due to its capability to reduce ID-irrelevant information, thereby facilitating better model generalization. However, CBMA fails to explicitly capture the interrelationship between the two modalities. Conversely, our proposed MFT module explores the relationship between cross attention and convolution, enabling the compensation of modality-specific information with minimal computational overhead.

\subsection{Parameter Analysis}

\begin{figure}[h]
	\centering
	\includegraphics[scale=0.3]{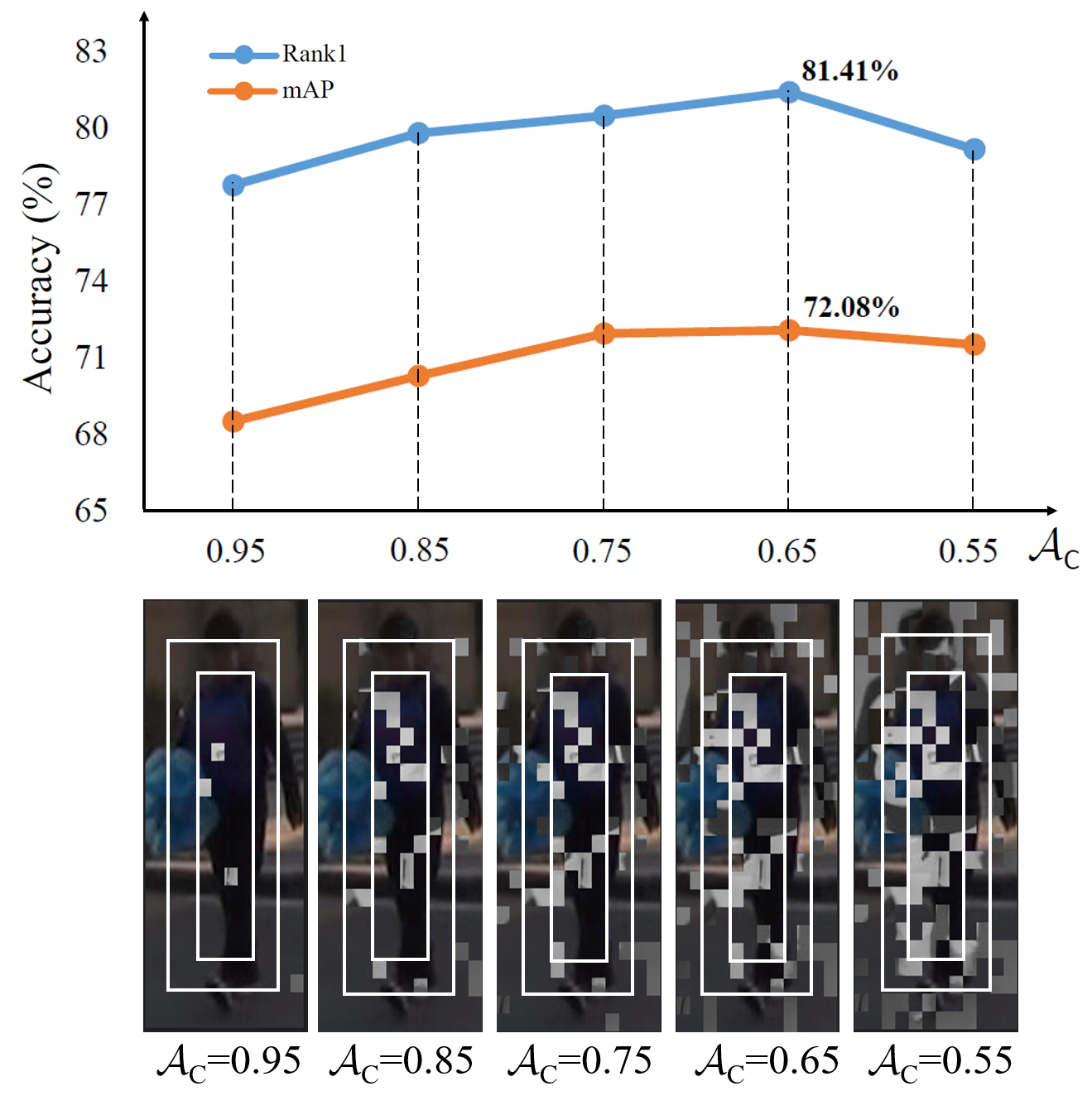}
	\centering
	\caption{Comparison with the different mix ratios in PedMix. The top diagram shows accuarcy for different mix ratios and the bottom one shows generated image. }
	\label{A}
\end{figure}

\begin{figure}[h]
	\centering
	\includegraphics[scale=0.3]{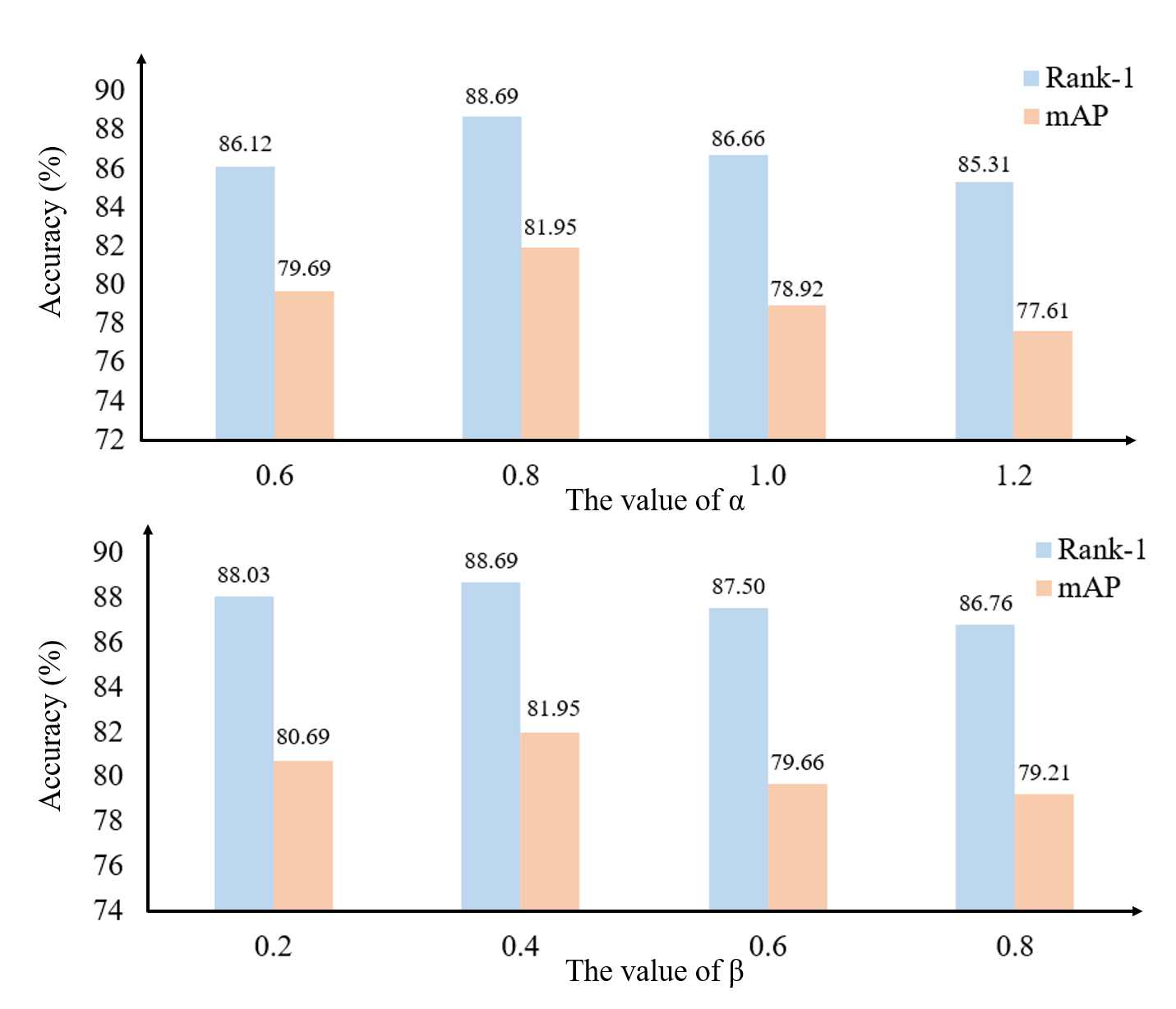}
	\centering
	\caption{Balance parameters for $\alpha$ and $\beta$ on RegDB dataset with Visible to Infrared mode. }
	\label{A}
\end{figure}

\begin{figure*}[h]
	\centering
	\includegraphics[scale=0.57]{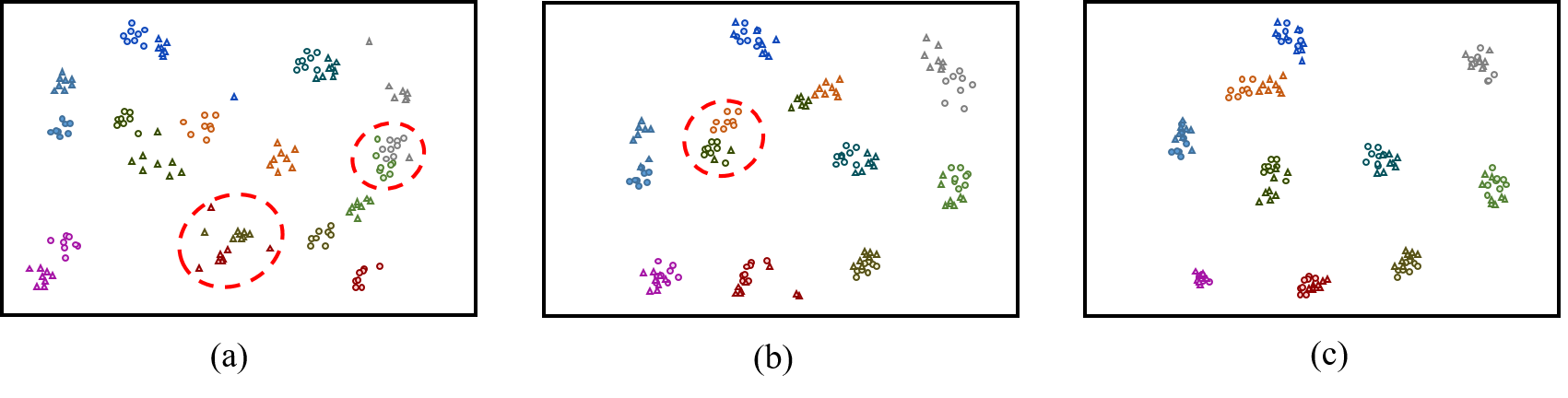}
	\caption{Feature distributions visualized by T-SNE. (a) Feature distribution of Baseline (B) method. (b) Feature distribution of B+PedMix method. (c) Feature distribution of B+PedMix+MFT method. Different colors represent different IDs, and the triangle and circle dots represent features extracted from infrared and visible modalities respectively.}
	\label{heatmap}
	\vspace{-3pt}
\end{figure*}

\noindent \textbf{Analysis of Mix Ratio.} The Rank-1 accuracy and mAP of different masking ratios $\mathcal{A}$ in PedMix are shown in Fig. 4. We empirically set the difference between $\mathcal{A}_C$, $\mathcal{A}_S$ and $\mathcal{A}_O$ to be 0.05. Our model is approximately robust to different masking ratios $\mathcal{A}$. The performance keeps improving before $\mathcal{A}_C$ arrives at 0.65  (i.e., $\mathcal{A}_S$ and $\mathcal{A}_O$ are set to 0.7 and 0.75). When $\mathcal{A}_C$ is greater than 0.65, the performance of the model decreases. We conclude that when $\mathcal{A}_C$ is set to 0.65, the images can be effectively mixed.

\begin{table}[htbp]
	\centering
	\caption{Comparison with different patch sizes.}
	\begin{tabular}{c|cccc}
		\hline\rule{0pt}{10pt}
		Patch Size & Rank-1 & Rank-10 & Rank-20 & mAP \\
		\hline\rule{0pt}{10pt}
		Baseline & 76.83 & 91.38 & 93.69 & 68.27 \\
		8×8   & 80.27 & 93.66 & 94.03 & 71.53 \\
		12×12 & 81.41 & 94.47 & 96.76 & 72.08 \\
		16×16 & 78.88 & 92.25 & 93.63 & 69.87 \\
		\hline
	\end{tabular}%
	\label{tab:addlabel}%
\end{table}%

\noindent \textbf{Analysis of Patch Size.} As shown in Table \uppercase\expandafter{\romannumeral7} our PedMix exhibits robustness across different patch sizes. We can observe that performance degrades when the patch size is chosen to be larger. This is because as the patch size increases, the boundaries between the center, sub-center, and outer regions become less distinct, introducing more contextual information. Conversely, if the patch size is too small, it can adversely affect the coherence of each region, thereby impacting performance.

\noindent \textbf{Analysis of Balance Parameters.} As shown in Fig. 5, we analyzed the impact of different balance parameters $\alpha$ and $\beta$ on the RegDB dataset with the Visible to Infrared mode. These balance parameters control the weights of the base loss $L_{base}$ and the modality feature loss $L_{MF}$. Initially, we set the value of $\beta$ to 0.4 and increase $\alpha$ from 0.6 to 1.2, obtaining the best setting value for $\alpha$. Similarly, with $\alpha$ fixed at 0.8, we increase $\beta$ from 0.2 to 0.8. Accordingly, we determine the optimal values to be $\alpha=0.8$ and $\beta=0.4$, achieving 88.69\% Rank-1 and 81.95\% mAP. Overall, our TMPA demonstrates robustness across different balance parameters. However, we observe that during the early stages of training, due to significant differences between modalities, the modality feature loss $L_{MF}$ can be excessively large, and high balance parameter $\beta$ may interfere with the model's convergence.

\subsection{Visualization}

\noindent \textbf{Visualization of Feature Distribution.} To intuitively understand the effectiveness of our proposed modules, we randomly selected feature representations of 10 identities and visualized their feature distributions via T-SNE\cite{van2008visualizing}. As shown in Fig. 6(a), when directly employing the Baseline (B) method, the intra-modality feature distribution is sparse and the cross modality features may get closely mapped together (encircled by the red dashed line). By using the proposed PedMix in Fig. 6(b), the intra-modality feature distribution is significantly drawn together. This is because the PedMix module aims to enhance modality-shared features, thereby facilitating a more compact alignment between two modalities. Moreover, when further employing the MFT module in Fig. 6(c), we notice that these extracted features have better distribution with compact intra-class distances and larger cross modality distances.

\begin{figure}[h]
	\centering
	\includegraphics[scale=0.085]{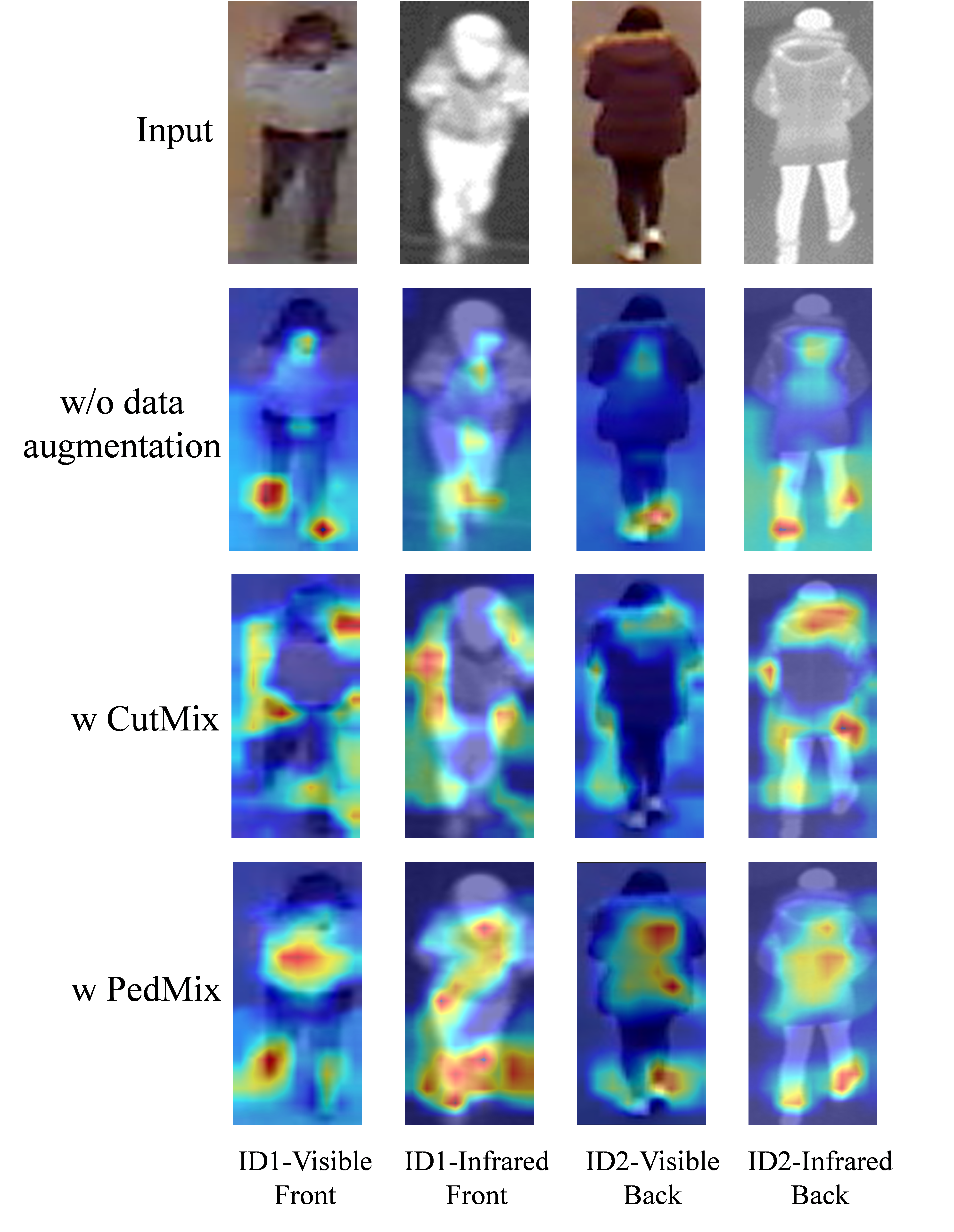}
	\caption{Heapmaps with Different Mix Strategies. The red parts represent high response while the blue parts represent low response.}
	\label{heatmap}
	\vspace{-10pt}
\end{figure}

\noindent \textbf{Visualization of Feature Heatmaps.} To intuitively observe the effectiveness of PedMix, we use four illurstrative pedestrian images from two different shooting views in two modalities to generate the heatmaps of different data augmentations by Grad-CAM \cite{selvaraju2017grad}. As shown in Fig. 7, without any data augmentation, the model focuses on a small region of the pedestrians (e.g., feet and heads). The CutMix augmentation, which mixes the images based on global information. The model's attention is generalized to the context region. Finally, with PedMix, the model focuses more on pedestrians, thereby boosting performance.

\begin{figure}[h]
	\centering
	\includegraphics[scale=0.35]{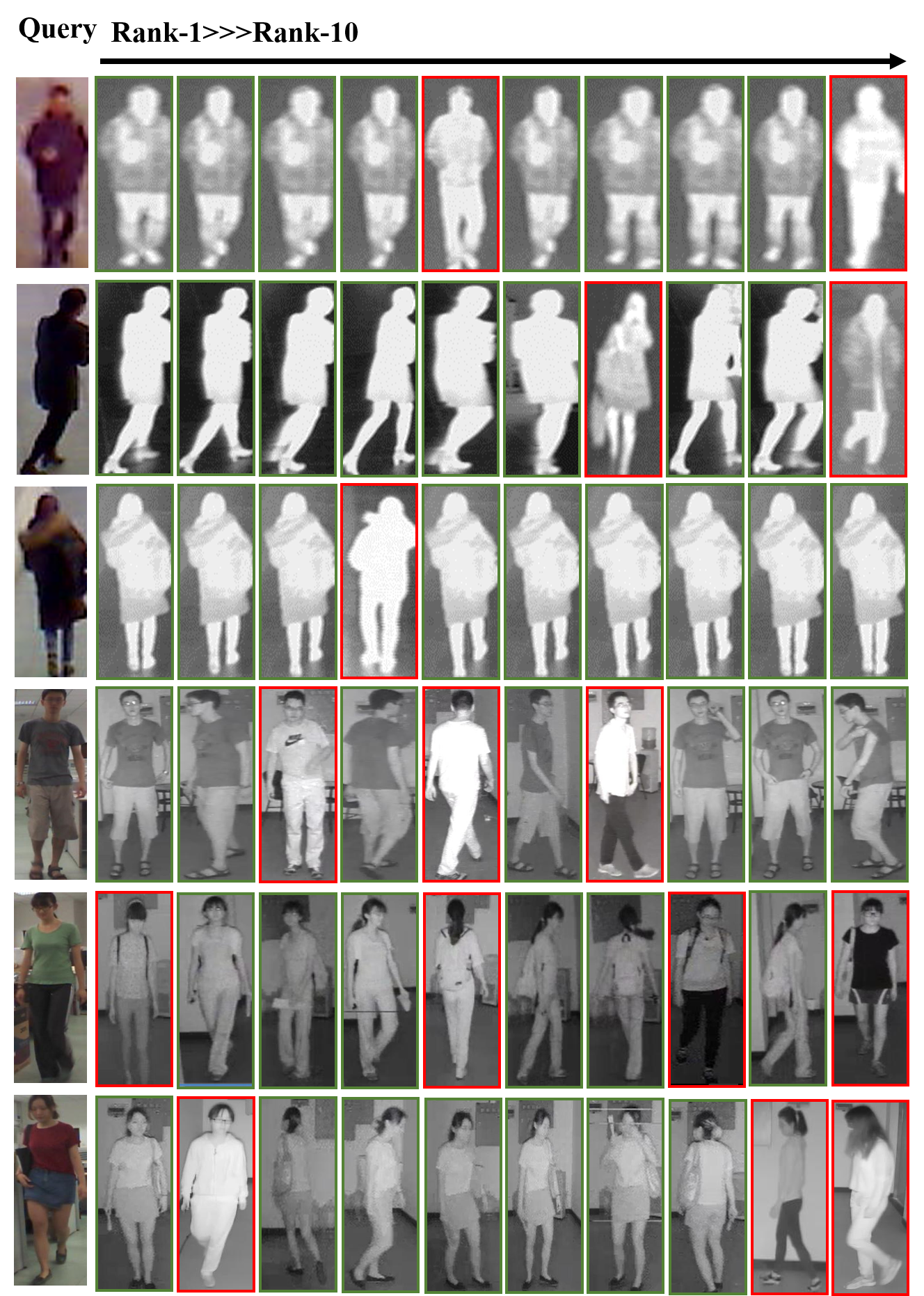}
	\caption{The top ten retrieved results of six randomly selected images from two datasets. The green box denotes correct matches and the red box denotes incorrect matches. (The first three images are from RegDB and the latter three are from SYSU-MM01, with the search mode set as Visible to Infrared.)}
	\label{heatmap}
	\vspace{-10pt}
\end{figure}

\noindent \textbf{Visualization of Retrieved Results.} As shown in Fig. 8, we randomly selected six pedestrian images from two datasets to validate the retrieval performance of our proposed TMPA. We observed that most top-ranked images are correct matches. However, there are some mismatches between the query and gallery images, primarily in the last ranks. In summary, our proposed TMPA demonstrates good performance on both datasets, effectively identifying cross-modal samples in the top ranks, which is helpful to the nighttime surveillance system.

\section{Conclusion}
In this paper, we propose a novel TMPA model for VI-ReID, aiming to leverage pedestrian-attentive features and learn modality-complete and -consistent representation. To this end, a region-based augmentation module PedMix is designed to enhance pedestrian region coherence. An efficient compensation module MFT is proposed to fully explore the discriminative modality-complete features while achieving less computational overhead by integrating cross attention and convolution. 

Extensive experiments are conducted on two commonly used VI-ReID datasets to demonstrate the effectiveness of our proposed TMPA model. The ablation studies and extensive analysis demonstrate the effectiveness of our proposed PedMix and MFT. Simultaneously, visual experiments allow us to observe that our proposed method reasonably generalizes the model to the entire pedestrian area, avoiding the introduction of ID-irrelevant information, and exhibits excellent retrieval performance.

\section{Acknowledgments}

This paper is partly supported by National Natural Science Foundation of China (No. 62276201) and the Fundamental Research Funds for the Central Universities (No. QTZX23025). The content is solely the responsibility of the authors and the authors have no relevant financial or non-financial interests to disclose.


\bibliographystyle{IEEEtran} 
\bibliography{bare_jrnl_new_sample4} 

\vfill

\end{document}